\documentclass[10pt,twocolumn,letterpaper]{article}
\usepackage{cvpr}
\usepackage{times}
\usepackage{epsfig}
\usepackage{graphicx}
\usepackage{amsmath}
\usepackage{amssymb}
\usepackage{multirow}
\usepackage{adjustbox}
\usepackage{siunitx}
\usepackage{subcaption}
\usepackage{textcomp}
\usepackage{xcolor}
\usepackage{authblk}
\usepackage{siunitx}
\DeclareMathOperator{\E}{\mathbb{E}}
\usepackage{amsmath}
\usepackage{bm}

\usepackage{booktabs}

\makeatletter
\renewcommand\AB@affilsepx{, \protect\Affilfont}
\makeatother

\usepackage[pagebackref=true,breaklinks=true,letterpaper=true,colorlinks,bookmarks=false]{hyperref}

\newif\ifarxiv
\arxivtrue 

\cvprfinalcopy 

\def\cvprPaperID{6179} 

\begin{document}

\title{Background Matting: The World is Your Green Screen}


%

\author{Soumyadip Sengupta}
\author{Vivek Jayaram}
\author{Brian Curless}
\author{Steve Seitz}
\author{Ira Kemelmacher-Shlizerman}

\affil{University of Washington}

\twocolumn[{%
\renewcommand\twocolumn[1][]{#1}%
\maketitle
\begin{center}
   \centering
   \vspace{-1em}
   \includegraphics[width=.98\textwidth]{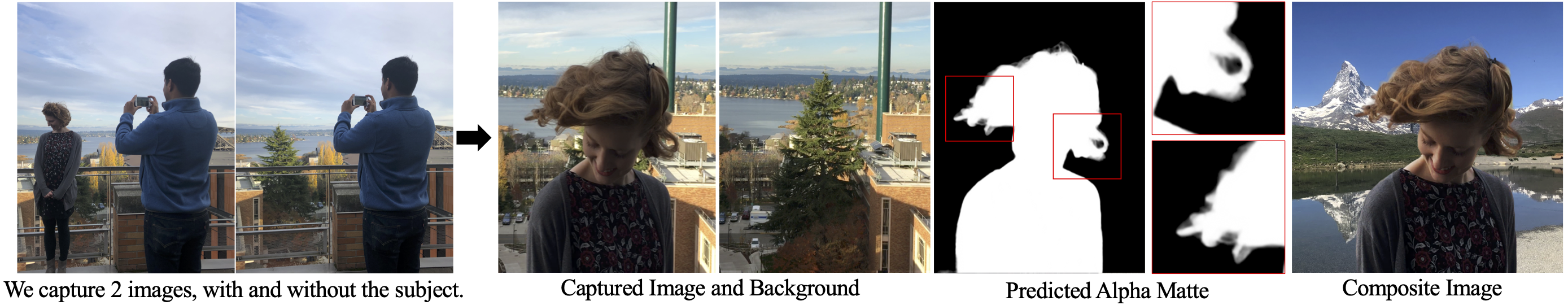}
   \captionof{figure}{\small Using a handheld smartphone camera, we capture two images of a scene, one with the subject and one without.  We employ a deep network with an adversarial loss to recover alpha matte and foreground color.  We composite the result onto a novel background.}
   \label{fig:teaser}
\end{center}%
}]

\begin{abstract}
\vspace{-0.5em}
We propose a method for creating a matte -- the per-pixel foreground color and alpha -- of a person by taking photos or videos in an everyday setting with a handheld camera.  Most existing matting methods require a green screen background or a manually created trimap to produce a good matte.  Automatic, trimap-free methods are appearing, but are not of comparable quality.  In our trimap free approach, we ask the user to take an additional photo of the background without the subject at the time of capture.  This step requires a small amount of foresight but is far less time-consuming than creating a trimap.  We train a deep network with an adversarial loss to predict the matte.  We first train a matting network with supervised loss on ground truth data with synthetic composites.  To bridge the domain gap to real imagery with no labeling, we  train another matting network guided by the first network and by a discriminator that judges the quality of composites. We demonstrate results on a wide variety of photos and videos and show significant improvement over the state of the art. 
\end{abstract}

\vspace{-1em}
\section{Introduction}
\label{sec:intro}


Imagine being able to easily create a matte — the per-pixel color and alpha — of a person by taking photos or videos in an everyday setting with just a handheld smartphone.  Today, the best methods for extracting (“pulling”) a good quality matte require either a green screen studio, or the manual creation of a {\em trimap} (foreground/background/unknown segmentation), a painstaking process that often requires careful painting around strands of hair.  Methods that require neither of these are beginning to appear, but they are not of comparable quality.  Instead, we propose taking an additional photo of the (static) background just before or after the subject is in frame, and using this photo to perform {\em background matting}.  Taking one extra photo in the moment requires a small amount of foresight, but the effort is tiny compared to creating a trimap after the fact. This advantage is even greater for video input.  Now, the world is your green screen.   

We focus on a method that is tuned to human subjects.
Still, even in this setting — pulling the matte of a person given a photo of the background — the problem is ill-posed and requires novel solutions. 

Consider the compositing equation for image $I$ given foreground $F$, background $B$, and mixing coefficient $\alpha$: $I=\alpha F + (1-\alpha) B.$
For color images and scalar $\alpha$, and given $B$, we have four unknowns ($F$ and $\alpha$), but only three observations per pixel ($I$).  Thus, the background matting problem is underconstrained.  Background/foreground differences 
provide a signal, but the signal is poor when parts of the person are similar in color to the background.  Furthermore, we do not generally have an image of the ideal background: the subject can cast shadows and cause reflections not seen in the photo taken without the subject, and exact, pixel-level alignment with no resampling artifacts between handheld capture of two photos is generally not attainable.  In effect, rather than the true $B$ that produced $I$, we have some perturbed version of it, $B'$. Finally, we can build on person segmentation algorithms to make the problem more tractable to identify what is semantically the foreground. However current methods, exhibit failures for complex body poses and fine features like hair and fingers. 

Given these challenges and recently published successes in solving matting problems, a deep learning approach is a natural solution.  We propose a deep network that estimates the foreground and alpha from input comprised of the original image, the background photo, and an automatically computed soft segmentation of the person in frame.  The network can also utilize several frames of video, useful for bursts or performance capture, when available. However, the majority of our results, including all comparisons to single-image methods, do not use any temporal cues.

We initially train our network on the Adobe Matting dataset~\cite{xu2017deep}, comprised of ground truth mattes that can be synthetically composited over a variety of backgrounds.  In practice, we found the domain gap between these synthetic composites and real-world images did not lead to good results using standard networks.  We partially close this gap in two ways: by augmenting the dataset and by devising a new network — a “Context Switching Block” — that more effectively selects among the input cues.  The resulting mattes for real images can still have significant artifacts, particularly evident when compositing onto a new background.  We thus additionally train the network in a self-supervised manner on real unlabelled input images using an adversarial loss to judge newly created composites and ultimately improve the matting process.

Our method has some limitations.  First, we do require two images.  Trimap-based methods arguably require two images as well for best results -- the trimap itself is a hand-made second image -- though they can be applied to any input photo.  Second, we require a static background and small camera motion; our method would not perform well on backgrounds with people walking through or with a camera that moves far from the background capture position.  Finally, our approach is specialized to foregrounds of (one or more) people.
That said, person matting without big camera movement in front of a static background is, we argue, a very useful and not uncommon scenario, and we deliver state-of-the-art results under these circumstances.


Our contributions include: $\bullet$ The first trimap-free automatic matting algorithm that utilizes a casually captured background. $\bullet$ A novel matting architecture (Context Switching Block) to select among input cues. $\bullet$ A self-supervised adversarial training to improve mattes on real images. $\bullet$ Experimental comparisons to a variety of competing methods on wide range of inputs (handheld, fixed-camera, indoor, outdoor), demonstrating the relative success of our approach. Our code and data is available at {\footnotesize \url{http://github.com/senguptaumd/Background-Matting}}.

\vspace{-0.5em}
\section{Related Work}
\label{sec:related}

\begin{figure*}[!ht]
	\centering
	\includegraphics[width=0.94\textwidth]{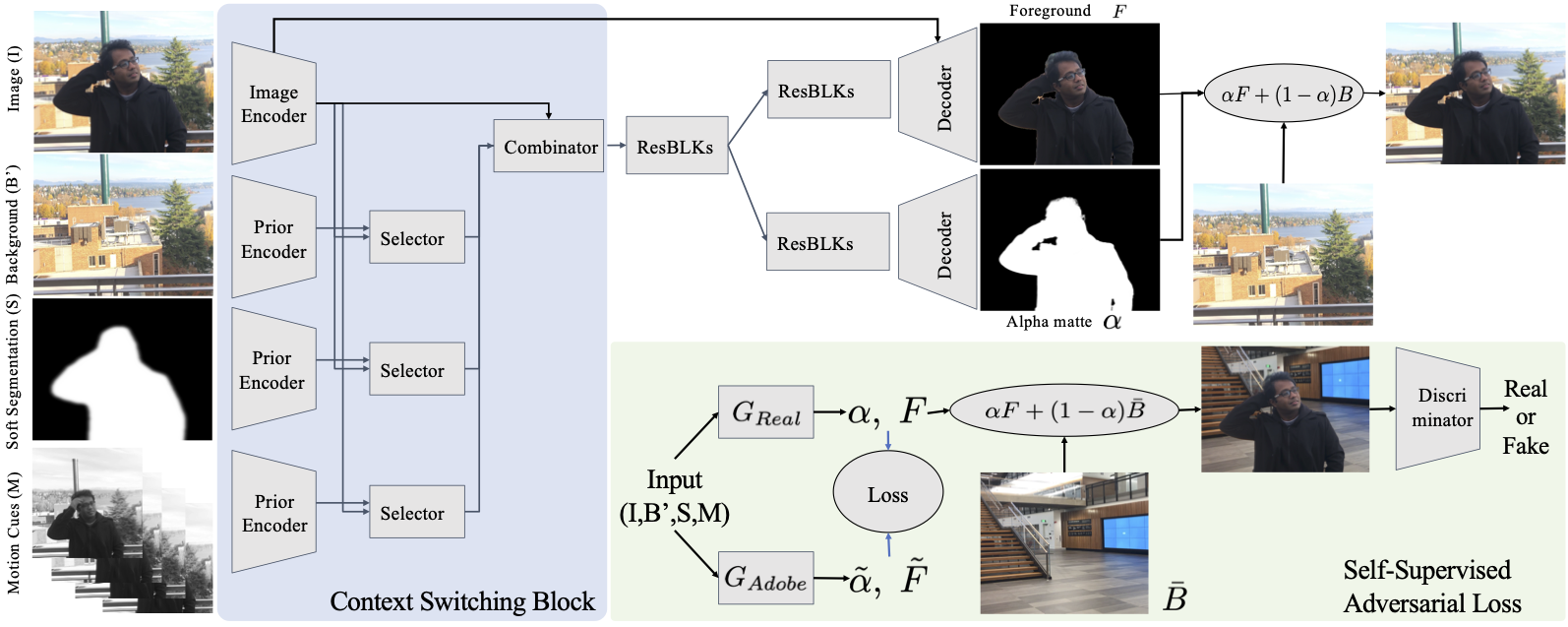}    \vspace{-.2cm}
	\caption{\small \textbf{Overview of our approach.} Given an input image $I$ and background image $B'$, we jointly estimate the alpha matte $\alpha$ and the foreground $F$ using soft segmentation $S$ and motion prior $M$ (for video only). We propose a Context Switching Block that efficiently combines all different cues. We also introduce self-supervised training on unlabelled real data by compositing into novel backgrounds.}
	\vspace{-1em}
	\label{fig:networks_main}
\end{figure*}

Matting is a standard technique used in photo editing and visual effects. In an uncontrolled setting, this is known as the ``natural image matting'' problem; pulling the matte requires solving for {\em seven} unknowns per pixel ($F, B, \alpha$) and is typically solved with the aid of a trimap.  In a studio, the subject is photographed in front of a uniformly lit, constant-colored background (e.g., a green screen); reasonable results are attainable if the subject avoids wearing colors that are similar to the background.  We take a middle ground in our work: we casually shoot the subject in a natural (non-studio) setting, but include an image of the background without the subject to make the matting problem more tractable.  In this section, we discuss related work on natural image matting, captured without unusual hardware.


\textbf{Traditional approaches.} Traditional (non-learning based) matting approaches generally require a trimap as input.  They can be roughly categorized into sampling-based techniques and propagation-based techniques. Sampling-based methods~\cite{gastal2010shared,chuang2001bayesian,he2011global,shahrian2013improving,wang2005iterative,wang2007optimized,aksoy2017designing} use sampling to build the color statistics of the known foreground and background, and then solve for the matte in the `unknown' region. Propagation-based approaches~\cite{chen2013knn,lee2011nonlocal,levin2007closed,levin2008spectral,sun2004poisson,grady2005random,he2010fast} aim to propagate the alpha matte from the foreground and the background region into the `unknown' region to solve the matting equation. Wang and Cohen~\cite{wang2008image} presents a nice survey of many different matting techniques.

\textbf{Learning-based approaches.} Deep learning approaches showed renewed success in natural image matting, especially in presence of user-generated trimaps. Some methods combine learning-based approaches with traditional techniques, e.g., KNN-matting~\cite{shen2016deep,cho2016natural}. Xu~\textit{et al.}~\cite{xu2017deep} created a matting dataset with real mattes and composited over a variety of backgrounds and trained a deep network to predict the alpha matte; these results were further improved by Lutz~\textit{et al.}~\cite{lutz2018alphagan} using an adversarial loss. Recently Tang~\textit{et al.}~\cite{Tang_2019_CVPR} proposed a hybrid of a sampling-based approach and learning to predict the alpha matte. Lu~\textit{et.al}~\cite{lu2019indices} proposed a new index-guided upsampling and unpooling operation that helps the network predict better alpha mattes. Cai~\textit{et al.}~\cite{cai2019disentangled} showed robustness to faulty user-defined trimaps. All of these methods only predict the alpha matte and not the foreground, leaving open the (non-trivial) problem of recovering foreground color needed for composites. 
Recently Hou~\textit{et al.}~\cite{hou2019context} introduced Context-Aware Matting (CAM) which simultaneously predicts the alpha and the foreground, thus solving the complete matting problem, but is not robust to faulty trimaps. In contrast to these methods (and the traditional approaches), our work jointly predicts alpha and foreground using an image of the background instead of a trimap.

Recently, researchers have developed algorithms that perform matting without a trimap, focusing mostly on humans (as we do). Aksoy~\textit{et.al.}~\cite{aksoy2018semantic} introduced fully automatic semantic soft segmentation for natural images. In \cite{zhu2017fast,shen2016deep} the authors perform portrait matting without trimap, utilizing segmentation cues. Trimap-free matting has also been extended to handle whole bodies in \cite{zhang2019late,chen2018semantic}. These methods aim to perform trimap prediction, followed by alpha prediction. Our work is also human-focused; we compare our approach with the recent state-of-the-art automatic human matting algorithm~\cite{zhang2019late} and obtain significantly better performance with the aid of the background image.

\textbf{Matting with known natural background.} Difference matting proposed by Qian and Sezan~\cite{qian1999video} attempts to solve matting with a natural background by simple background subtraction and thresholding but is very sensitive to the threshold and produces binary mattes. Similarly, change detection via background subtraction~\cite{piccardi2004background,elgammal2000non} generally does not produce alpha mattes with foreground and considers shadows to be part of the foreground.  Some traditional approaches like Bayesian matting~\cite{chuang2001bayesian} and Poisson matting~\cite{sun2004poisson,gong2009near} can handle known background in their framework, but additionally require trimaps.  

\textbf{Video Matting.} 
Researchers have also focused on video-specific methods. Chuang \textit{et.al.}~\cite{chuang2002video} extended Bayesian Matting to videos by utilizing the known background and optical flow, requiring trimaps for  keyframes. Flow-based temporal smoothing can be used~\cite{lee2010temporally,shahrian2014temporally} (again with trimaps) to encourage temporal coherence. 




\vspace{-0.5em}
\section{Our Approach}
\label{sec:model}
The input to our system is an image or video of a person in front of a static, natural background, plus an image of just the background.  The imaging process is easy, just requiring the user to step out of the frame after the shot to capture the background, and works with any camera with a setting to lock the exposure and focus (e.g., a smartphone camera).  For handheld capture, we assume camera motion is small and align the background to a given input image with a homography.  From the input, we also extract a soft segmentation of the subject.  For video input, we can additionally utilize nearby frames to aid in matting.

At the core of our approach is a deep matting network $G$ that extracts foreground color and alpha for a given input frame, augmented with background, soft segmentation, and (optionally nearby video frames), and a discriminator network $D$ that guides the training to generate realistic results.  In Section~\ref{sec:sup_train}, we describe the matting network, which contains a novel architecture -- a ``Context-switching block'' -- that can combine different input cues selectively.  We first train a copy of this network $G_{\rm Adobe}$ with supervision using the Adobe Matting Dataset~\cite{xu2017deep}. We use known foreground and alpha mattes of non-transparent objects, which are then composited over a variety of backgrounds (i.e., real source images, but synthetic composites).  Our matting network, along with some data augmentation, help overcome some of the domain gap between the synthetically composited imagery and real data that we later capture with a consumer camera (e.g., a smartphone).

In Section~\ref{sec:self_sup_train}, we describe a self-supervised scheme to bridge the domain gap further and to generally improve the matting quality.  The method employs an adversarial network comprised of a separate copy of the deep matting network, $G_{\rm Real}$, that tries to produce a matte similar to the output of $G_{\rm Adobe}$ and a discriminator network $D$ that scores the result of compositing onto a novel background as real or fake.  We train $G_{\rm Real}$ and $D$ jointly on real inputs, with supervision provided by (the now fixed) $G_{\rm Adobe}$ network applied to the same data.

\subsection{Supervised Training on the Adobe Dataset}
\label{sec:sup_train}
Here we describe our deep matting network, which we first train on the Adobe Matting Dataset, restricted to the subset of non-transparent objects.  The network takes as input an image $I$ with a person in the foreground, an image of the background $B'$ registered to $I$ (as noted earlier, $B'$ is not the same as the true $B$ with subject present), a soft segmentation of the person $S$, and (optionally for video) a stack of temporally nearby frames $M$, and produces as output a foreground image $F$ and alpha matte $\alpha$.  To generate $S$, we apply person segmentation~\cite{deeplabv3plus2018} and then erode (5 steps), dilate (10 steps), and apply a Gaussian blur ($\sigma=5$).  When video is available, we set $M$ to be the concatenation of the two frames before and after $I$, i.e., $\{I_{-2T},I_{-T},I_{+T},I_{+2T}\}$ for frame interval $T$; these images are converted to grayscale to ignore color cues and focus more on motion cues. In the absence of video, we simply set $M$ to $\{I,I,I,I\}$, also converted to grayscale.  We denote the input set as $X \equiv \{I,B',S,M\}$. The network with weight parameters $\theta$ thus computes:
\vspace{-0.5em}
\begin{equation}
\label{eq:setup}
(F, \alpha)  = G(X;\theta).
\vspace{-0.5em}
\end{equation}

In designing and training the network, the domain gap between the Adobe dataset and our real data has proven to be a significant driver in our choices as we describe below.

A natural choice for $G$ would be a residual-block-based encoder-decoder~\cite{zhu2017unpaired} operating on a concatenation of the inputs $\{I,B',S,M\}$.  Though we would expect such a network to learn which cues to trust at each pixel when recovering the matte, we found that such a network did not perform well.  When training on the Adobe synthetic-composite data and then testing on real data, the resulting network tended to make errors like trusting the background $B'$ too much and generating holes whenever $F$ was too close in color; the network was not able to bridge the domain gap.

Instead, we propose a new Context Switching block (CS block) network (Figure~\ref{fig:networks_main}) to combine features more effectively from all cues, conditioned on the input image. 
When, e.g., a portion of the person matches the background, the network should focus more on  segmentation cue in that region. The network has four different encoders for $I$, $B'$, $S$, and $M$ that separately produce 256 channels of feature maps for each. It then combines the image features from $I$ with each of $B'$, $S$ and $M$ separately by applying 1x1 convolution, BatchNorm, and ReLU (`Selector' block in Fig.~\ref{fig:networks_main}), producing 64-channel features for each of the three pairs. Finally, these three 64-channel features are combined with the original 256-channel image features with 1x1 convolution, BatchNorm, and ReLU (the `Combinator' block in Fig.~\ref{fig:networks_main}) to produce encoded features which are passed on to the rest of the network, consisting of residual blocks and decoders. We observe that the CS Block architecture helps to generalize from the synthetic-composite Adobe dataset to real data (Figure~\ref{fig:cs_block}). More network architecture details are provided in the supplementary material.

We train the network with the Adobe Matting dataset~\cite{xu2017deep} which provides 450 ground truth foreground image $F^*$ and alpha matte $\alpha^*$ (manually extracted from natural images).  We select the subset of 280 images corresponding to non-transparent objects (omitting, e.g., objects made of glass).  As in~\cite{xu2017deep}, we can compose these foregrounds over known backgrounds drawn from the MS-COCO dataset, augmented with random crops of varying resolutions, re-scalings, and horizontal flips.  These known backgrounds $B$ would not be the same as captured backgrounds $B'$ in a real setting.  Rather than carefully simulate how $B$ and $B'$ might differ, we simply perturbed $B$ to avoid training the network to rely too much on its exact values.  In particular, we generated each $B'$ by randomly applying either a small gamma correction $\gamma \sim \mathcal{N}(1,0.12)$ to $B$ or adding gaussian noise $\eta \sim \mathcal{N}(\mu \in [-7,7],\sigma \in [2,6])$ around the foreground region. Further, to simulate imperfect segmentation guidance $S$ we threshold the alpha matte and then erode (10-20 steps), dilate (15-30 steps) and blur ($\sigma \in [3,5,7]$) the result. For the motion cue $M$, we applied random affine transformations to foreground+alpha before compositing onto the background, followed by conversion to grayscale.  To compute $I$ and $M$ we used the compositing equation with $B$ as the background, but we provided $B'$ as the input background to the network.

Finally, we train our network $G_{\rm Adobe} \equiv G(\cdot;\theta_{\rm Adobe})$ on the Adobe dataset with supervised loss:
\vspace{-2mm}
\begin{equation}
\label{eq:syn}
\begin{split}
&\min_{\theta_{\rm Adobe}} E_{X \sim p_{X}} [\|\alpha-\alpha^*\|_1 + \|\nabla (\alpha)-\nabla (\alpha^*)\|_1  \\
 &+ 2\|F-F^*\|_1  + \| I - \alpha F - (1-\alpha) B\|_1],
\end{split}
\end{equation}
where $(F,\alpha) = G(X;\theta_{\rm Adobe})$, and the gradient term on $\alpha$ encourages sharper alpha mattes~\cite{zhang2019late}. 


\subsection{Adversarial Training on Unlabelled Real data}
\label{sec:self_sup_train}
Although our proposed Context Switch block (CS block) combined with  data augmentation significantly helps in bridging the gap between real images and synthetic composites created with the Adobe dataset, it still fails to handle all difficulties present in real data. Theses difficulties include (1)~traces of background around fingers, arms, and hairs being copied into the matte; (2)~segmentation failing; (3)~significant parts of the foreground color matching the background color; (4)~misalignment between the image and the background (we assume only small misalignment). To handle these cases, we aim to learn from unlabelled, real data (real images + backgrounds) with self-supervision.

The key insight is that significant errors in the estimated matte typically result in unrealistic composites over novel backgrounds.  For example, a bad matte might contain a chunk of the source background, which, when composited over a new background, will have a piece of the original background copied over the new background, a major visual artifact.  Thus, we can train an adversarial discriminator to distinguish between fake composites and (already captured) real images to improve the matting network.

The matting network ($G_{\rm Real} \equiv G(\cdot;\theta_{\rm Real})$) and discriminator  network $D$ can be trained end-to-end based on just a standard discriminator loss.  However, $G_{\rm Real}$ could settle on setting $\alpha = 1$ everywhere, which would result in simply copying the entire input image into the composite passed to $D$.  This solution is ``optimal'' for $G_{\rm Real}$, since the input image is indeed real and should fool $D$.  Initializing with $G_{\rm Adobe}$ and fine-tuning with a low learning rate (was necessary for stable training with a discriminator) is not very effective. It does not allow significant changes to network weights needed to generate good mattes on real data.

Instead, we use $G_{\rm Adobe}$ for teacher-student learning. In particular, for a real training image $I$ and associated inputs comprising $X$, we obtain $(\tilde{F}, \tilde{\alpha}) = G(X;\theta_{\rm Adobe})$ to serve as ``pseudo ground-truth''. We can now train with an adversarial loss {\it and} a loss on the output of the matting network $G(X;\theta_{\rm Real})$ when compared to ``pseudo ground-truth'', following~\cite{sfsnetSengupta18}; this second loss is given small weight which is reduced between epochs during training.  Though we initialize $\theta_{\rm Real}$ in the standard randomized way, the network is still encouraged to stay similar to the behavior of $G_{\rm Adobe}$ while having the flexibility to make significant changes that improve the quality of the mattes. We hypothesize that this formulation helps the network to avoid getting stuck in the local minimum of $G_{\rm Adobe}$, instead finding a better minimum nearby for real data. 

We use the LS-GAN~\cite{mao2017least} framework to train our generator $G_{\rm Real}$ and discriminator $D$. For the generator update we minimize:
\vspace{-1mm}
\begin{equation}
\label{eq:gen}
\begin{split}
\min_{\theta_{\rm Real}} 
&\E_{X,\bar{B} \sim p_{X,\bar{B}}} 
[(D(\alpha F + (1-\alpha) \bar{B})-1)^2 \\
&+\lambda \{2\|\alpha-\tilde{\alpha}\|_1 + 4 \|\nabla (\alpha)-\nabla (\tilde{\alpha})\|_1   \\
&+ \|F-\tilde{F}\|_1  + \| I - \alpha F - (1-\alpha) B'\|_1 \}],
\end{split}
\end{equation}
where $(F,\alpha) = G(X;\theta_{\rm Real})$, $\bar{B}$ is a given background for generating a composite seen by $D$, and we set $\lambda$ to 0.05 and reduce by $1/2$ every two epochs during training to allow the discriminator to play a significant role. We use a higher weight on the alpha losses (relative to Equation~\ref{eq:syn}), especially the gradient term to encourage sharpness. 

For the discriminator, we minimize:
\vspace{-1mm}
\begin{equation}
\label{eq:dis}
\begin{split}
\min_{\theta_{\rm Disc}} \E_{X,\bar{B} \sim p_{X,\bar{B}}} 
&[(D(\alpha F + (1-\alpha) \bar{B}))^2] \\ 
&+ \E_{I\in p_{data}} [(D(I)-1)^2],
\end{split}
\end{equation}
where $\theta_{\rm Disc}$ represents the weights of the discriminator network and again $(F,\alpha) = G(X;\theta_{\rm Real})$. 

As a post-process, we threshold the matte at $\alpha > 0.05$, extract the largest $N$ connected components, and set $\alpha=0$ for pixels not in those components, where $N$ is the number of disjoint person segmentations in the image.

\vspace{-0.5em}
\section{Experimental Evaluation}

\begin{figure*}
	\centering
	\includegraphics[width=.9\textwidth]{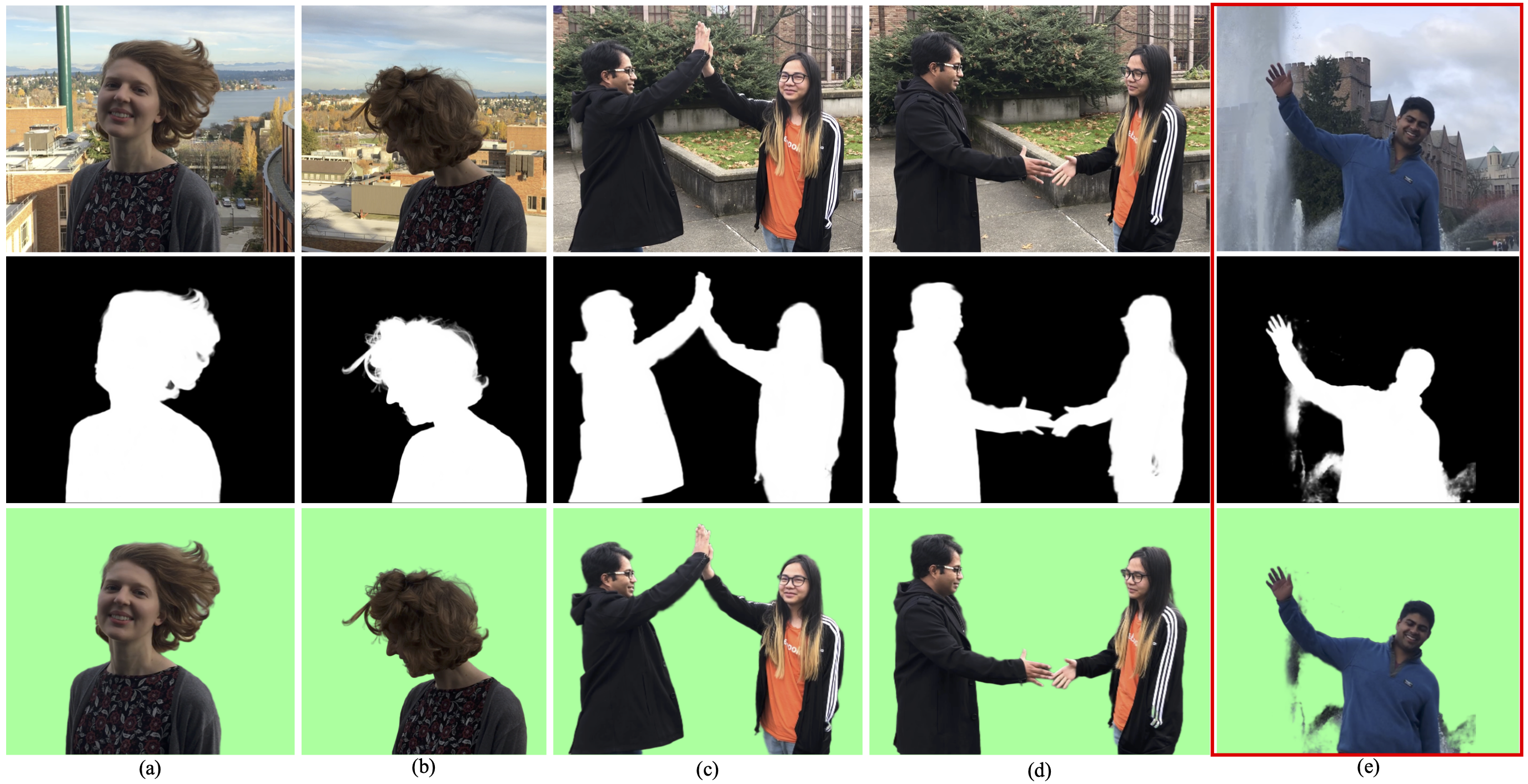}    \vspace{-.2cm}
	\caption{\small (a-e) Resulted alphas and foregrounds for photos taken with handheld camera against natural backgrounds; (e) is an example failure case with dynamic background (fountain). See video results in the supplementary.} 
	\vspace{-1em}
	\label{fig:our_result}
\end{figure*}
\label{sec:baseline}

We compared our approach with a variety of alternative methods, esp. recent deep matting algorithms that have performed well on benchmarks: \textbf{BM}: Bayesian Matting~\cite{chuang2001bayesian} - traditional, trimap-based method that can accept a known background~\cite{chuang2002video}.  (An alternative, Poisson Matting~\cite{sun2004poisson,gong2009near} with known background, performed much worse.). \textbf{CAM}: Context-Aware Matting~\cite{hou2019context} - trimap-based deep matting technique that predicts both alpha and foreground. \textbf{IM}: Index Matting~\cite{lu2019indices} - trimap-based deep matting technique that predicts only alpha. \textbf{LFM}: Late Fusion Matting~\cite{zhang2019late} - trimap-free deep matting algorithm that predicts only alpha.

\subsection{Results on Synthetic-Composite Adobe Dataset}
\label{sec:syn}
We train $G_{\rm Adobe}$ on 26.9k exemplars: 269 objects composited over 100 random backgrounds, plus perturbed versions of the backgrounds as input to the network. We train with batch-size 4, learning rate $1e^{-4}$ with Adam optimizer. 

\begin{table}
\vspace{-0.5em}
\setlength\tabcolsep{2pt}
	\centering
	\small
	\captionsetup{justification=centering}
		\begin{tabular}{cccc}
			\toprule
			 Algorithm & Additional Inputs & SAD & MSE($10^{-2}$) \\ 
			\midrule
			\textbf{BM} & Trimap-10, $B$ & 2.53& 1.33\\
			\textbf{BM} & Trimap-20, $B$ & 2.86& 1.13\\
			\textbf{BM} & Trimap-20, $B'$ & 4.02& 2.26\\
			\midrule
			\textbf{CAM} & Trimap-10 & 3.67& 4.50\\
			\textbf{CAM} & Trimap-20 & 4.72& 4.49\\
			\midrule
			\textbf{IM} & Trimap-10 & 1.92& 1.16\\
			\textbf{IM} & Trimap-20 & 2.36& 1.10\\
			\midrule
			\textbf{Ours-Adobe} & $B$ & \textbf{1.72}& \textbf{0.97}\\
			\textbf{Ours-Adobe} & $B'$ & \textbf{1.73}& \textbf{0.99}\\
			\bottomrule
		\end{tabular}
		\caption{\small Alpha matte error on Adobe Dataset (lower is better).}
	\vspace{-1.5em}
	\label{tab:adobe}
\end{table}

We compare results across 220 synthetic composites from the Adobe Dataset~\cite{xu2017deep}: 11 held-out mattes of human subjects composed over 20 random backgrounds, in Table~\ref{tab:adobe}. We computed a trimap for each matte through a process of alpha matte thresholding and dilation as described in~\cite{xu2017deep}. We dilated by 10 and 20 steps to generate two different trimaps (more steps gives wider unknown region).  We additionally computed a perturbed background $B'$ by applying small random affine transformation (translate $\in \mathcal{N}(0,3)$, rotate $\in \mathcal{N}(0,1.3^{\circ})$ and small scaling and shear) followed by gamma correction $\gamma \sim \mathcal{N}(1,0.12)$ and gaussian noise $\eta \sim \mathcal{N}(\mu \in [-5,5],\sigma \in [2,4])$.  For our approach, we only evaluated the result of applying the $G_{\rm Adobe}$ network (`Ours-Adobe'), since it was trained only on the Adobe data, as were the other learning-based approaches we compare to. 
We rescaled all images to $512\times512$ and measure the SAD and MSE error between the estimated and ground truth (GT) alpha mattes, supplying algorithms with the two different trimaps and with backgrounds $B$ and $B'$ as needed.  We omitted LFM from this comparison, as the released model was trained on additional training data, along with the training set of Adobe dataset. That said, it produces a SAD and MSE of 2.00, 1.08$e^{-2}$, resp., while our method achieves error of 1.72, 0.97$e^{-2}$.


We observe that our approach is more robust to background perturbation when compared to BM, and it improves on all other trimap-based matting algorithms (BM, CAM, IM). As trimaps get tighter, the trimap-based matting algorithms get better, but tight trimaps are time-consuming to create in practice. The goal of our work is to fully eliminate the need for manually created trimaps. 



\subsection{Results on Real Data}
\label{sec:real}

We captured a mix of handheld and fixed-camera videos, taken indoors and outside using a smartphone (iPhone 8).  The fixed-camera setup consisted of an inexpensive selfie stick tripod. In each case, we took a video with the subject moving around, plus a shot of the background (single video frame) with no subject.  All frames were captured in HD (1920$\times$1080), after which they were cropped to 512$\times$512 (input resolution to our network) around the segmentation mask for one person or multiple. We retrain $G_{\rm Adobe}$ on 280k composites consisting of 280 objects from Adobe Dataset~\cite{xu2017deep}. We then train separate copies of $G_{\rm Real}$, one each on handheld videos and fixed camera videos, to allow the networks to focus better on the input style. For handheld videos we account for small camera shake by  aligning the captured background to individuals frames through homography.  In total, we trained on 18k frames for hand-held camera and 19k frames for fixed camera. We captured 3390 additional background frames for $\bar{B}$. We use a batch-size of 8, learning rate of $1e^{-4}$ for $G_{\rm Real}$ and $1e^{-5}$ for $D$ and update $D$ with Adam optimizer. We also update the weights of $D$ after 5 successive updates of $G_{\rm Real}$.

\begin{table}[!h]
\vspace{-0.5em}
\setlength\tabcolsep{2pt}
	\centering
\small
	\captionsetup{justification=centering}
    	\vspace{-.5em}
		\begin{tabular}{c|ccccc}
			\toprule
			 Ours vs. &  much better & better & similar & worse & much worse \\ 
			 \midrule
			 BM & 52.9\% & 41.4\% & 5.7\% & 0\% & 0\%\\
			 CAM & 30.8\% & 42.5\% & 22.5\% & 4.2\% & 0\%\\
			 IM & 26.7\% & 55.0\% & 15.0\% & 2.5\% & 0.8\%\\
			 LFM & 72.0\% & 20.0\% & 4.0\% & 3.0\% & 1\%\\
			\bottomrule
		\end{tabular}
		\vspace{-.5em}
		\caption{\small User study on 10 real world videos (fixed camera).}
	\vspace{-1.5em}
	\label{tab:real-fixed}
\end{table}

\begin{table}[!h]
\setlength\tabcolsep{2pt}
	\centering
	\small
	\captionsetup{justification=centering}
    	\vspace{-.5em}
		\begin{tabular}{c|ccccc}
			\toprule
			 Ours vs. &  much better & better & similar & worse & much worse \\ 
			 \midrule
			 BM & 61.0\% & 31.0\% & 3.0\% & 4.0\% & 1.0\% \\
			 CAM & 43.3\% & 37.5\% & 5.0\% & 4.2\% & 10.0\%\\
			 IM & 33.3\% & 47.5\% & 5.9\% & 7.5\% & 5.8\%\\
			 LFM & 65.7\% & 27.1\% & 4.3\% & 0\% & 2.9\%\\
			\bottomrule
		\end{tabular}
		\vspace{-.5em}
		\caption{\small User study on 10 real world videos (handheld).}
	\vspace{-1em}
	\label{tab:real-hand}
\end{table}

To compare algorithms on real data, we used 10 handheld videos and 10 fixed-camera videos as our (held-out) test data.  The BM, CAM, and IM methods each require trimaps.  We did not manually create trimaps (esp. for video sequences which is infeasible).  Instead, we applied segmentation~\cite{deeplabv3plus2018}, and labeled each pixel with person-class probability $>0.95$ as foreground, $<0.05$ as background, and the rest as unknown. We tried alternative methods, including background subtraction, but they did not work as well.


To evaluate results, we could not compare numerically to ground truth mattes, as none were available for our data.  Instead, we composited the mattes over a green background and performed a user study on the resulting videos. Since IM and LFM do not estimate $F$ (needed for compositing), we set $F = I$ for these methods.  We also tried estimating $F$ directly from the matting equation (given $\alpha$ and $B'$), but the results were worse (see supplementary material). We do not use any temporal information and set $M=\{I,I,I,I\}$ for all comparisons to prior methods.



\begin{figure}[ht!]
	\centering
	\includegraphics[width=0.42\textwidth]{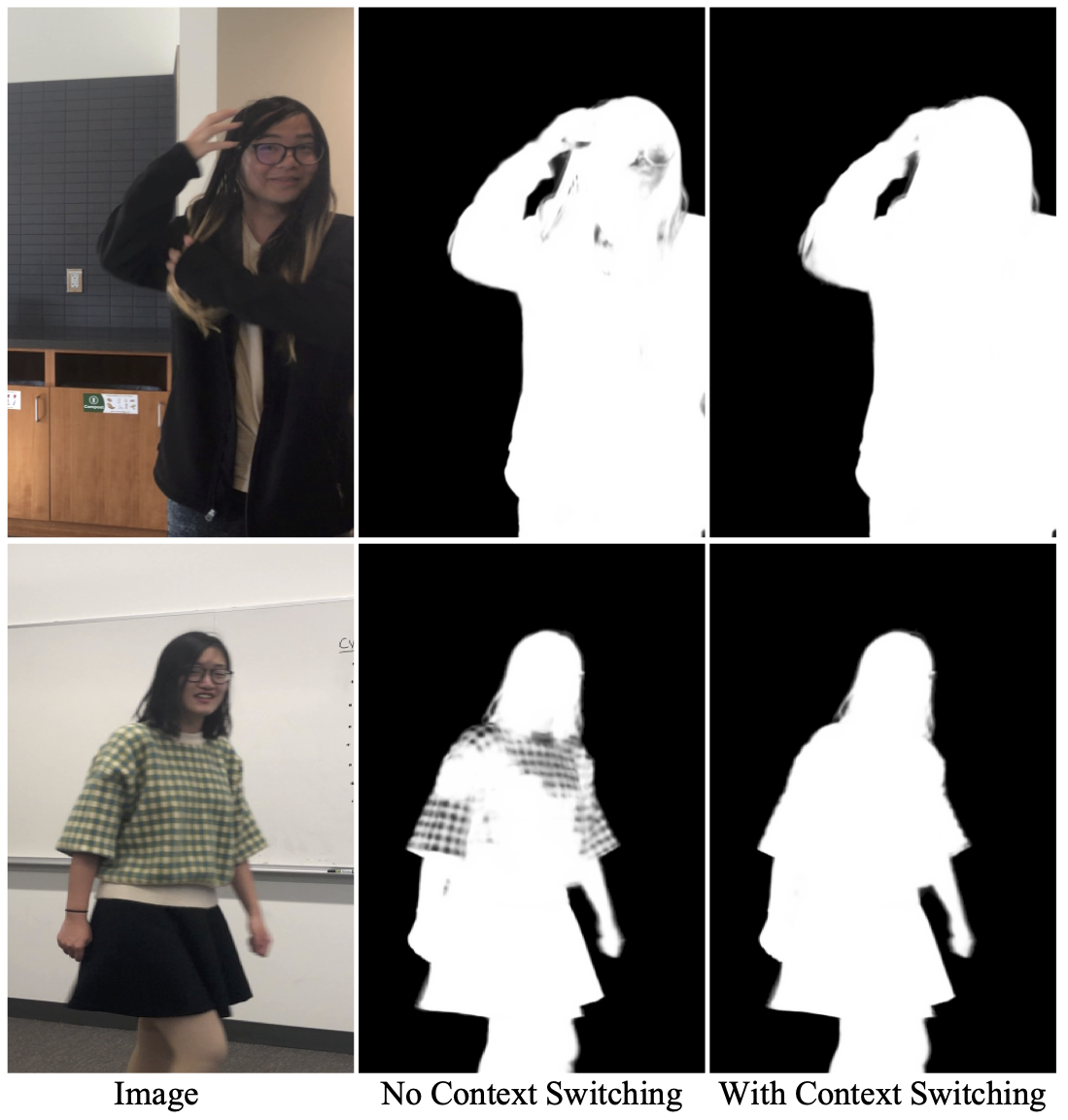}    \vspace{-0.5em}
	\caption{\small Role of Context Switching Block (CS Block).} 
	\vspace{-0.5em}
	\label{fig:cs_block}
\end{figure}

\begin{figure}[ht!]
	\centering
	\includegraphics[width=0.42\textwidth]{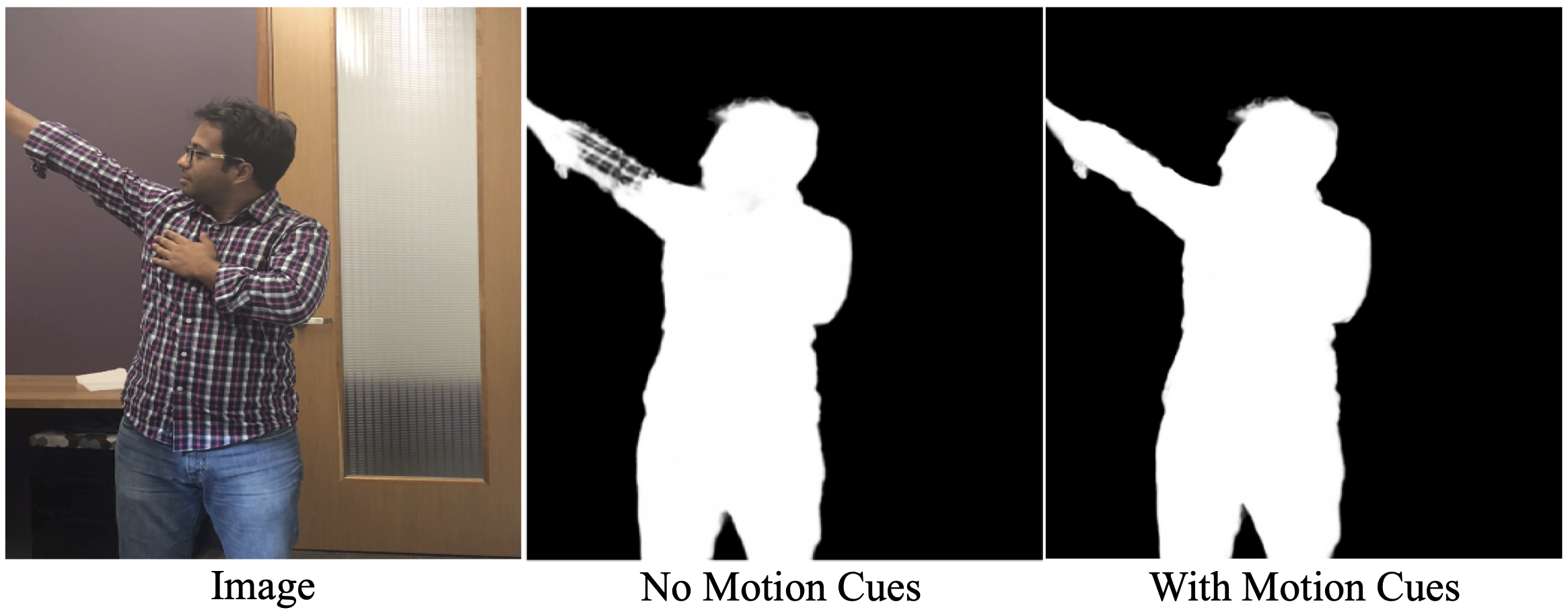}    \vspace{-0.5em}
	\caption{\small Role of motion cues.}
	\vspace{-1em}
	\label{fig:motion_cues}
\end{figure}


In the user study, we compared the composite videos produced by $G_{\rm Real}$ network (`Ours-Real') head-to-head with each of the competing algorithms.  Each user was presented with a web page showing the original video, our composite, and a competing composite; the order of the last two was random.  The user was then asked to rate composite A relative to B on a scale of 1-5 (1 being `much worse', 5 `much better').  Each video pair was rated $\sim$ 10 users.

\begin{figure*}[!ht]
	\centering
	\includegraphics[width=0.95\textwidth]{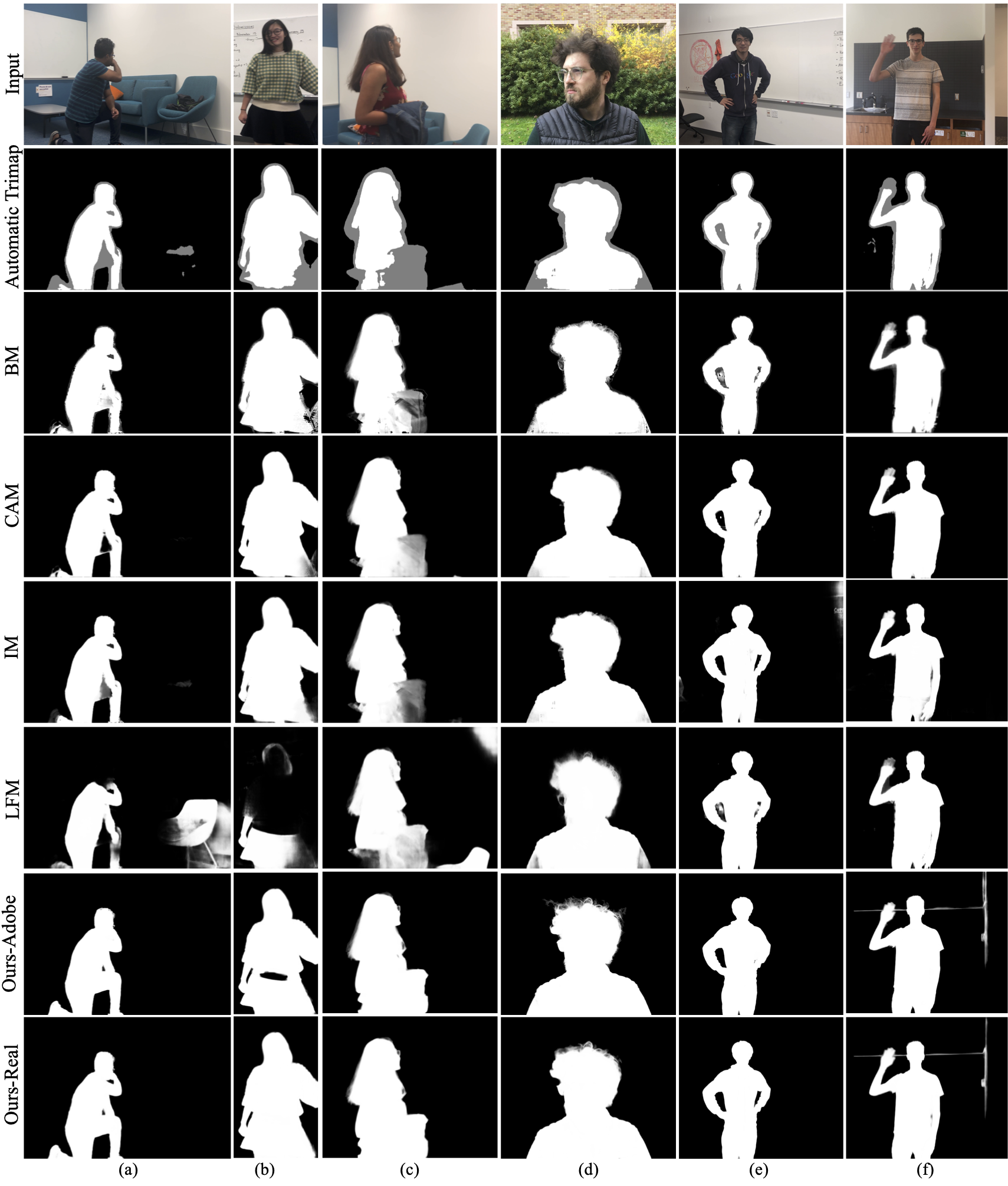}    \vspace{-.2cm}
	\caption{\small Comparison of matting methods with camera fixed (a,b,c) and handheld (d,e,f). Our method fails in (f) due to misregistration.} 
	\vspace{-1em}
	\label{fig:qual_image}
\end{figure*}

The results of the user study, with scores aggregated over all test videos, are shown in Tables~\ref{tab:real-fixed} and \ref{tab:real-hand}. Overall, our method significantly outperformed the alternatives.  The gains of our method are somewhat higher for fixed-camera results; with handheld results, registration errors can still lead to matting errors due to, e.g., parallax in non-planar background scenes (see Fig~\ref{fig:qual_image}(f)).

Single image results are shown in Figure~\ref{fig:qual_image}, again demonstrating improvement of our method over alternatives.  We note that LFM in particular has difficulty zeroing in on the person.  More results generated by our approach with handheld camera in natural backgrounds are shown in Figure~\ref{fig:our_result}. In (c), (d) we show examples of multiple people interacting in a single image, and in (e) we show a failure case with a dynamic background, the fountain. \textbf{Please see supplementary material for video results and more image results.}




\vspace{-0.5em}
\section{Ablation Studies}
\label{sec:abla}
\textbf{Role of motion cues.} As shown in Figure~\ref{fig:motion_cues}, video motion cues $M$  can help in predicting a cleaner matte when foreground color matches the background.  (Note:  we did not use motion cues when comparing to other methods, regardless of input source.)
\begin{table}[!h]
\setlength\tabcolsep{2pt}
	\centering
	\small
	\captionsetup{justification=centering}
    	\vspace{-.5em}
		\begin{tabular}{c|ccccc}
			\toprule
			  &  much better & better & similar & worse & much worse \\ 
			 \midrule
			 handheld & 16.4\% & 35.5\% & 42.7\% & 5.4\% & 0\%\\
			 fixed-camera & 17.3\% & 15.5\% & 51.8\% & 10\% & 5.4\%\\
			\bottomrule
		\end{tabular}
		\vspace{-.5em}
		\caption{\small User Study: Ours-Real vs Ours-Adobe.}
	\vspace{-.5em}
	\label{tab:syn-real}
\end{table}

\textbf{`Ours-Real' vs `Ours-Adobe'.} As expected, `Ours-Adobe' outperformed `Ours-Real' on the synthetic-composite Adobe dataset on which `Ours-Adobe' was trained. `Ours-Real' achieved a SAD score of 3.50 in comparison to 1.73 of `Ours-Adobe'. However `Ours-Real' significantly outperformed `Ours-Adobe' on real data as shown by qualitative examples in Figure~\ref{fig:qual_image} and by an additional user study (Table \ref{tab:syn-real}).  The gain of `Ours-Real' in the user study ($\sim$ 10 users per pair-wise comparison) was larger for handheld captures; we suspect this is because it was trained with examples having alignment errors.  (We did try training `Ours-Adobe' with alignment errors introduced into $B'$ but found the results degraded overall.) 


\textbf{Role of Context Switching Block (CS Block).} We compare our CS Block architecture to a standard residual-block-based encoder-decoder~\cite{zhu2017unpaired} scheme that was run on a naive concatenation of $I$, $B'$, $S$, and $M$.  We find that the concatenation-based network learns to focus too much on color difference between $I$ and $B'$ and generates holes when their colors are similar.  The CS Block architecture effectively utilizes both segmentation and color difference cues, along with motion cues when present, to produce better matte, as shown in Figure \ref{fig:cs_block} (more in supplementary). Empirically, we observe that the CS block helps significantly in 9 out of 50 real videos, especially when foreground color is similar to the background.

\section{Conclusion}
\vspace{-0.5em}
\label{sec:discussion}
We have proposed a  background matting technique that enables casual capture of high quality foreground+alpha mattes in natural settings.  Our method  requires the photographer to take a shot with a (human) subject and without, not moving much between shots.  This approach avoids using a green screen or painstakingly constructing a detailed trimap as typically needed for high matting quality. A key challenge is the absence of real ground truth data for the background matting problem. We have developed a deep learning framework trained on synthetic-composite data and then adapted to real data using an adversarial network. 

\textbf{Acknowledgements.} This work was supported by NSF/Intel Visual and Experimental Computing Award \#1538618 and the UW Reality Lab.




\clearpage
\newpage




\newpage
{\small
\bibliographystyle{ieee_fullname}
\bibliography{ref}
}

\ifarxiv
    \newpage
    \clearpage
    \appendix
    \section{Overview}
We provide additional details and results in this Appendix. In Sec. \ref{sec:arch} we describe the details of our network architecture. In Sec.~\ref{sec:auto_tri} we clarify our choice of automatic trimap generation, especially to show why background subtraction is not good enough for this problem.  In Sec.~\ref{sec:fg}, we show why algorithms that do not predict foreground $F$ introduce additional artifacts in compositing. In Sec.~\ref{sec:abla_sup}, we provide additional qualitative examples for ablation studies. Specifically, we show the role of Content Switching Block in Sec. \ref{sec:cs_block}, motion cue in Sec. \ref{sec:motion_cue},  self-supervised adversarial training on real data in Sec. \ref{sec:real_vs_syn} and robustness w.r.t. segmentation and background in Sec. \ref{sec:back_seg}.

\section{Network Architectures}
\label{sec:arch}

Our proposed matting network $G_{\rm Adobe}$, shown again for reference in Figure~\ref{fig:networks_main} (same as Figure 2 in the main paper) consists of the Context Switching Block (CS Block) followed by residual blocks (ResBLKs) and decoders to predict alpha matte $\alpha$ and foreground layer $F$. Below we describe the network architecture in details. Most of our Generator architecture, especially residual blocks and decoders, and Discriminator architecture are based on that of \cite{zhu2017unpaired}.
\\
\noindent
\textbf{Generator $G_{\rm Adobe}$\\}
\noindent
\textbf{`Image Encoder' and `Prior Encoder' (CS Block)}: C64(k7) - C*128(k3) - C*256(k3)\\
`CN(kS)' denotes convolution layers with N $S \times S$ filters with stride 1, followed by Batch Normalization and ReLU. `C*N(kS)' denotes convolution layers with N $S \times S$ filters with stride 2, followed by Batch Normalization and ReLU. The output of `Image Encoder' layer produces a blob of spatial resolution $256 \times 128 \times 128$. All convolution layers do not have any bias term.\\
\noindent
\textbf{`Selector' (CS Block)}: C64(k1)\\
`CN(kS)' denotes convolution layers with N $S \times S$ filters with stride 1, followed by Batch Normalization and ReLU. The `Selector' block takes as input the concatenation of image feature and a prior feature as a blob of spatial resolution $512 \times 128 \times 128$. The output of the `Selector' network is a blob of spatial resolution $64 \times 128 \times 128$. The goal of the `Selector' block is to generate prior features conditioned on the image. This will help the network to generalize from synthetic-composite dataset (on which it was trained) to real images by not over-relying on one kind of features (e.g. color difference with background).\\
\noindent
\textbf{`Combinator' (CS Block)}: C256(k1)\\
`CN(kS)' denotes convolution layers with N $S \times S$ filters with stride 1, followed by Batch Normalization and ReLU. The `Combinator' block takes as input the concatenation of image features of spatial resolution $256 \times 128 \times 128$, along with 3 other prior features from the `Selector' network of spatial resolution $64 \times 128 \times 128$ each. Thus the input to the `Combinator' block is of spatial resolution $448 \times 128 \times 128$ and the output is of spatial resolution $256 \times 128 \times 128$. The `Combinator' block learns to combine the individual priors features with the original image feature for the goal of improving matting.\\

\noindent
\textbf{`ResBLKs'}: K ResBLK \\
The output of the combinator is first passed through $K=7$ `ResBLK's and then provided as input to 2 separate `ResBLK's of $K=3$ for alpha matte and foreground layer separately. All `ResBLK's operate at a spatial resolution of $256 \times 128 \times 128$. Each `ResBLK' consists of Conv256(k3) - BN -ReLU - Conv256(k3) - BN, where `BN' denote Batch Normalization.\\

\noindent
\textbf{`Decoder' for alpha matte $\alpha$:} CU*128(k3)-CU*64(k3)-Co1(k7)-Tanh\\
The input to the `Decoder' is of resolution $256 \times 128 \times 128$.`CU*N(kS)' denotes bilinear up-sampling by factor of 2, followed by convolution layer with $N$ $S \times S$ filters with stride 1, Batch Normalization and ReLU. The last layer Co3k(7) consists of only convolution layers of 1 $7\times7$ filters, followed by Tanh layer. This scales the output alpha between $(-1,1)$.\\

\noindent
\textbf{`Decoder' for foreground $F$:} CU*128(k3)-CU*64(k3)-Co1(k7)\\
The input to the `Decoder' is of resolution $256 \times 128 \times 128$.`CU*N(kS)' denotes bilinear up-sampling by factor of 2, followed by convolution layer with $N$ $S \times S$ filters with stride 1, Batch Normalization and ReLU. There is also a skip connection from the image input features of resolution $128 \times 256 \times 256$ which is combined with the output of CU*128(k3) and passed on to CU*64(k3). The last layer Co3k(7) consists of only convolution layers of 1 $7\times7$ filters.\\

\noindent
\textbf{Discriminator $D$: } C*64(k4) - C*I128(k4) - C*I256(k4) - C*I512(k4)\\
We use 70 × 70 PatchGAN \cite{isola2017image}. `C*N(kS)' denotes convolution layers with N $S \times S$ filters with stride 2, followed by leaky ReLUs of slope 0.2. `I' denotes the presense of Instance Norm before leaky ReLU, in all layers except the first one. After the last layer a convolution filter of kernel $4 \times 4$ is applied to produce a 1 dimensional output. The PatchGAN is applied over the whole composite image by convolving with every $70 \times 70$ patch to determine if it is real or fake.

\begin{figure*}[!ht]
	\centering
	\includegraphics[width=0.98\textwidth]{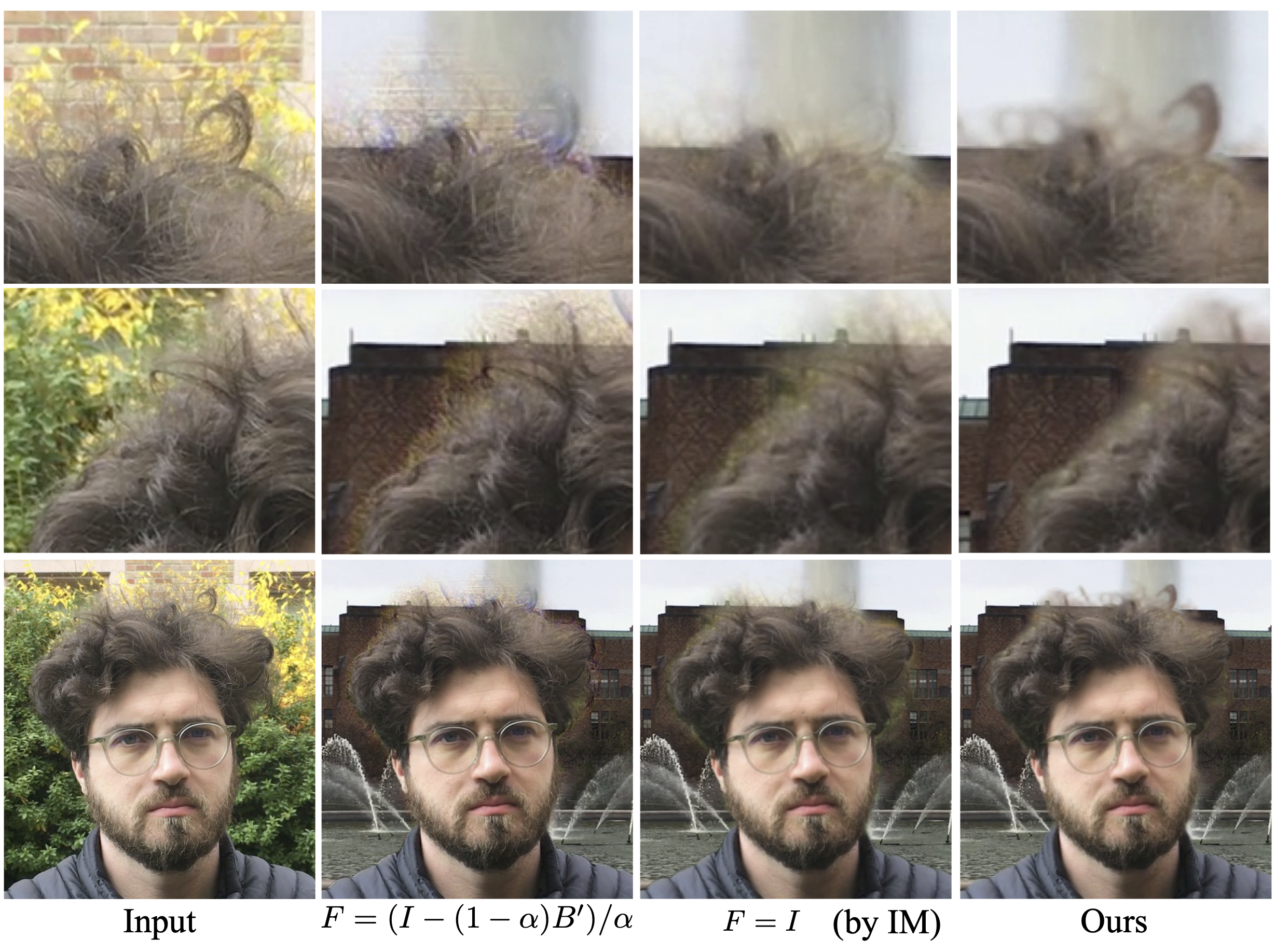}    
	\caption{\small \textbf{Choice of Foreground layer.} For baseline algorithms, IM and LFM, that do not predict the foreground layer $F$, we observe that $F=I$ produces less visible artifacts compared to predicting $F$ from the matting equation using the captured background $B'$. Notice how some of the brick texture creeps into the foreground when solving for $F$ with the matting equation.  We also show that our approach, which jointly estimates $F$ and $\alpha$, produces less artifacts in compositing.} 
	\label{fig:choiceF}
\end{figure*}

\section{Experimental Evaluation}

\subsection{Automatic Trimap Generation}
\label{sec:auto_tri}
We compare our method with algorithms that require user defined trimaps (CAM, IM, BM). It is extremely time consuming to annotate trimaps for every frame of a video, or even for a bunch of keyframes whose trimaps then need to be propagated to the remaining frames and then touched up. As described in the paper, we instead created trimaps automatically by applying segmentation~\cite{deeplabv3plus2018}, and labelling each pixel with person-class probability $>0.95$ as foreground, $<0.05$ as background, and the rest as unknown. We tried, and rejected, alternative methods, including background subtraction and erosion-dilateion of the segmentation mask, which we now describe and illustrate here for completeness.

Background subtraction is popularly used for change detection, but is extremely sensitive to color differences and produces \textit{only} a binary matte. Thus it is not in itself a suitable candidate for matting. However background subtraction could in principle be used to generate a trimap. In our experiments, we observed that the best thresholds varied from image to image and even then produced mediocre results (see hand-tuned example in Figure~\ref{fig:trimap}). We also tried erosion-dilation of the segmentation mask (erode 5 steps, dilate 15 steps) to try to produce a fixed width `unknown' band but we often ended up with mattes that pulled in the background as a part of the foreground. Figure~\ref{fig:trimap} shows examples of trimaps and resulting alpha mattes using erosion-dilation, hand-tuned thresholding of background subtraction, and our probability-thresholded `Automatic Trimap' method.

\begin{figure}[!ht]
	\centering
	\includegraphics[width=0.48\textwidth]{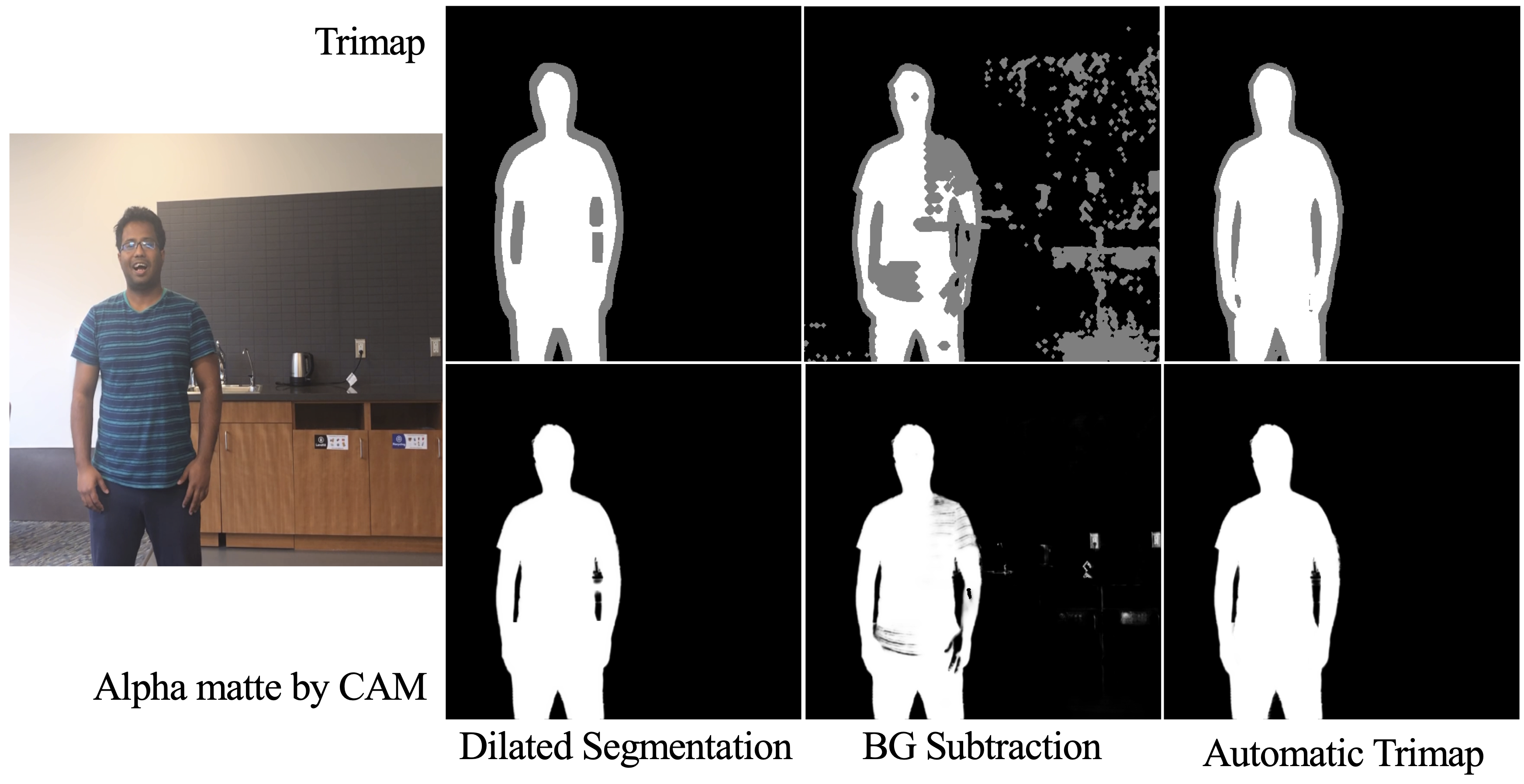}    
	\caption{\small \textbf{Automatic Trimap generation.} Our choice of automatic trimap generation from the probability estimates of the segmentation network performs better than background subtraction or erosion-dilation of the segmentation mask.} 
	\label{fig:trimap}
\end{figure}

\begin{figure*}[!ht]
	\centering
	\includegraphics[width=0.98\textwidth]{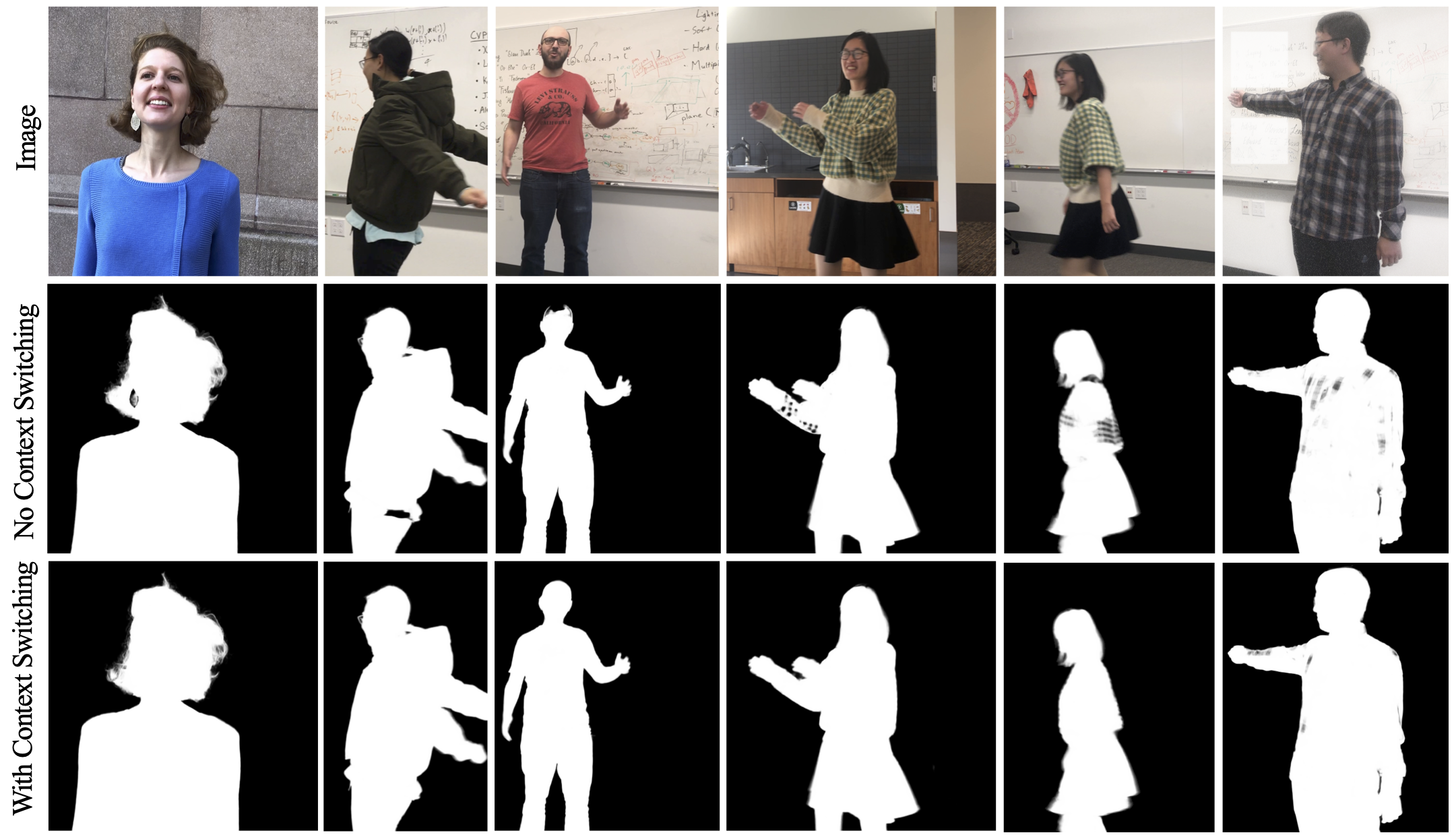}
	\caption{\small \textbf{Role of CS Block.} When foreground color coincides with the background color, Context Switching Block utilizes soft segmentation to predict the correct matte. `No Context Switching' produces holes when foreground color matches strongly with the background.}
	\label{fig:cs_block_ab}	
\end{figure*}

\subsection{Predicting Foreground layer $F$}
\label{sec:fg}
To produce composites -- the primary reason for extracting a matte in the first place -- we require both $\alpha$ and $F$.  Since IM and LFM do not estimate $F$, we are still left with the task of estimating it ourselves.  Why is this non-trivial?  From the matting equation ($I = \alpha F + (1-\alpha) B$) after estimating only $\alpha$, we can say that observed pixel color $I$ must be $\alpha$ of the way along a line segment from $B$ to $F$.  Clearly, there is an infinite family of $B$'s and $F$'s that satisfy this constraint; thus, given just $I$ and $\alpha$, we cannot readily infer $F$.  Our seemingly naive solution is to set $F = I$ for these methods.  We also tried estimating $F$ directly from the matting equation given $B'$ -- i.e., $F = (I-(1-\alpha)B')/\alpha$ when $\alpha \neq 0$, with $F$ clamped so that each color channel is in $[0,1]$ -- but the results were worse than $F=I$, largely due to discrepancies between $B'$ and $B$, particularly in the handheld camera case, where small misalignments can arise. We show a comparison of these two options for IM matting, plus a comparison to our result, in Figure~\ref{fig:choiceF}.  The figure shows that matting-equation based estimation of $F$ pulls some of the background texture into the matte.  The $F=I$ solution is better, but picks up some of the background colors (a bit of green and yellow in this case), since it is just copying foreground-background mixed pixels and blending them over another background.  Our method, which estimates $F$ directly, shows fewer artifacts.  Our matte is a bit softer, but it captures structure like the curls on the top of the head; further, some of the apparent sharpness of the other composites comes from copying over too much of the original image rather (which has detail) rather than fully separating $F$ from the background.


\subsection{Results on Real Data}

After conducting the user study, we realized that we had given a slight advantage to our method by retaining only the largest $\alpha > 0$ connected component for our background matting approach but not for the competing approaches.  We noted that the CAM and IM methods had small ``floaters'' after matting, so applied the connected component removal to these videos and re-ran the study.  We did not observe that BM had floaters in our examples; any that appeared to remain, were actually connected to the largest component by ``bridges'' of small $\alpha$.  LFM was more problematic. We found that LFM would at times pull in pieces of the background that were larger than the foreground person; the result of retaining the largest connected component would then mean losing the foreground subject altogether, an extremely objectionable artifact.  Rather than continuing to refine the post-process for LFM, we simply did not apply a post-process for its results.  As seen in the videos, LFM, in any case, had quite a few other artifacts that made it not competitive with the others.

Table~\ref{tab:real-fixed1} and \ref{tab:real-hand1} shows the result of the updated user study. We observe that the results are similar to the ones reported in the main paper (i.e., Tables~2 and~3 in the main paper); the excess connected components for CAM and IM did not have a significant impact relative to other errors in matte estimation. 


\begin{table}[!h]
\setlength\tabcolsep{2pt}
	\centering
\small
	\captionsetup{justification=centering}
		\begin{tabular}{c|ccccc}
			\toprule
			 Ours vs. &  much better & better & similar & worse & much worse \\ 
			 \midrule
			 BM & 52.9\% & 41.4\% & 5.7\% & 0\% & 0\%\\
			 CAM & 40.8\% & 36.7\% & 19.2\% & 3.3\% & 0\%\\
			 IM & 25.8\% & 52.5\% & 18.4\% & 2.5\% & 0.8\%\\
			 LFM & 72.0\% & 20.0\% & 4.0\% & 3.0\% & 1\%\\
			\bottomrule
		\end{tabular}
		\vspace{0.5em}
		\caption{\small User study on 10 real world videos (fixed camera).}
	\vspace{-.5em}
	\label{tab:real-fixed1}
\end{table}



\begin{table}[!h]
\setlength\tabcolsep{2pt}
	\centering
	\small
	\captionsetup{justification=centering}
		\begin{tabular}{c|ccccc}
			\toprule
			 Ours vs. &  much better & better & similar & worse & much worse \\ 
			 \midrule
			 BM & 61.0\% & 31.0\% & 3.0\% & 4.0\% & 1.0\% \\
			 CAM & 45.0\% & 35.0\% & 5.0\% & 5.0\% & 10.0\%\\
			 IM & 34.2\% & 46.6\% & 6.7\% & 2.5\% & 10.0\%\\
			 LFM & 65.7\% & 27.1\% & 4.3\% & 0\% & 2.9\%\\
			\bottomrule
		\end{tabular}
		\vspace{0.5em}
		\caption{\small User study on 10 real world videos (handheld).}
	\label{tab:real-hand1}
\end{table}

\section{Ablation Study}
\label{sec:abla_sup}

In this section, we provide additional details and more results for the ablation studies already presented in the main paper. Specifically, we analyze (i) Role of Context Switching Block (ii) Role of motion cues and (iii) Compare `Ours-Real' to `Ours-Adobe'.

\subsection{Role of Context Switching Block}
\label{sec:cs_block}
Here we go in more depth on the paper's CS block ablation study, again showing that the CS Block network is largely effective in utilizing all cues and in generalizing better from the synthetic-composite Adobe dataset~\cite{xu2017deep} to real world images. To this end we train $G_{\rm Adobe}$ with CS Block on the Adobe dataset which we term as `Ours-Adobe' (with Context Switching). Additionally, we construct another network $G_{\rm concat}$ (No Context Switching), where we remove the CS block. The input to this network is the concatenation of the input image, background, soft segmentation, and motion cues $\{I,B',S,M\}$ , which is passes through the `Image Encoder' architecture to produce a $256 \times 128 \times 128$ dimensional feature. Since there is no CS  Block, we directly pass this feature to the ResBLKs and then continue through the same architecture presented in Figure \ref{fig:networks_main}. We also train this network on the Adobe dataset, following the exact protocol of training `Ours-Adobe'.

We then test both $G_{\rm Adobe}$ and $G_{\rm concat}$ on our real video dataset. Note that for this experiment we use the motion cue $M=\{I_{-2T},I_{-T},I_{+T},I_{+2T}\}$ for both $G_{\rm Adobe}$ and $G_{\rm concat}$. We captured videos at 60fps and set $T=20$ frames.

In Figure~\ref{fig:cs_block_ab} of this document and in Figure~4 of the main paper, we show multiple examples from different videos where a part of the foreground person matches the background color. $G_{\rm concat}$ (`No Context Switching') fails in these situations, this is because while training on the synthetic-composite dataset it learns to focus too much on the color differences to perform matting and fails when colors coincide. On the other hand $G_{\rm Adobe}$ (`With Context Switching') handles color coincidences better by utilizing the soft segmentation cues provided in the input. Thus the CS Block learns to properly utilize the cues at hand, when compared to `No Context Switching', which focuses more on the color differences to produce a matte. Note that, the holes produced by `No Context Switching' as shown in Figure~\ref{fig:cs_block_ab} appears in multiple frames of that video where the color coincides significantly; we show only 1 sample from all of these frames.

\begin{figure}[!ht]
	\centering
	\includegraphics[width=0.48\textwidth]{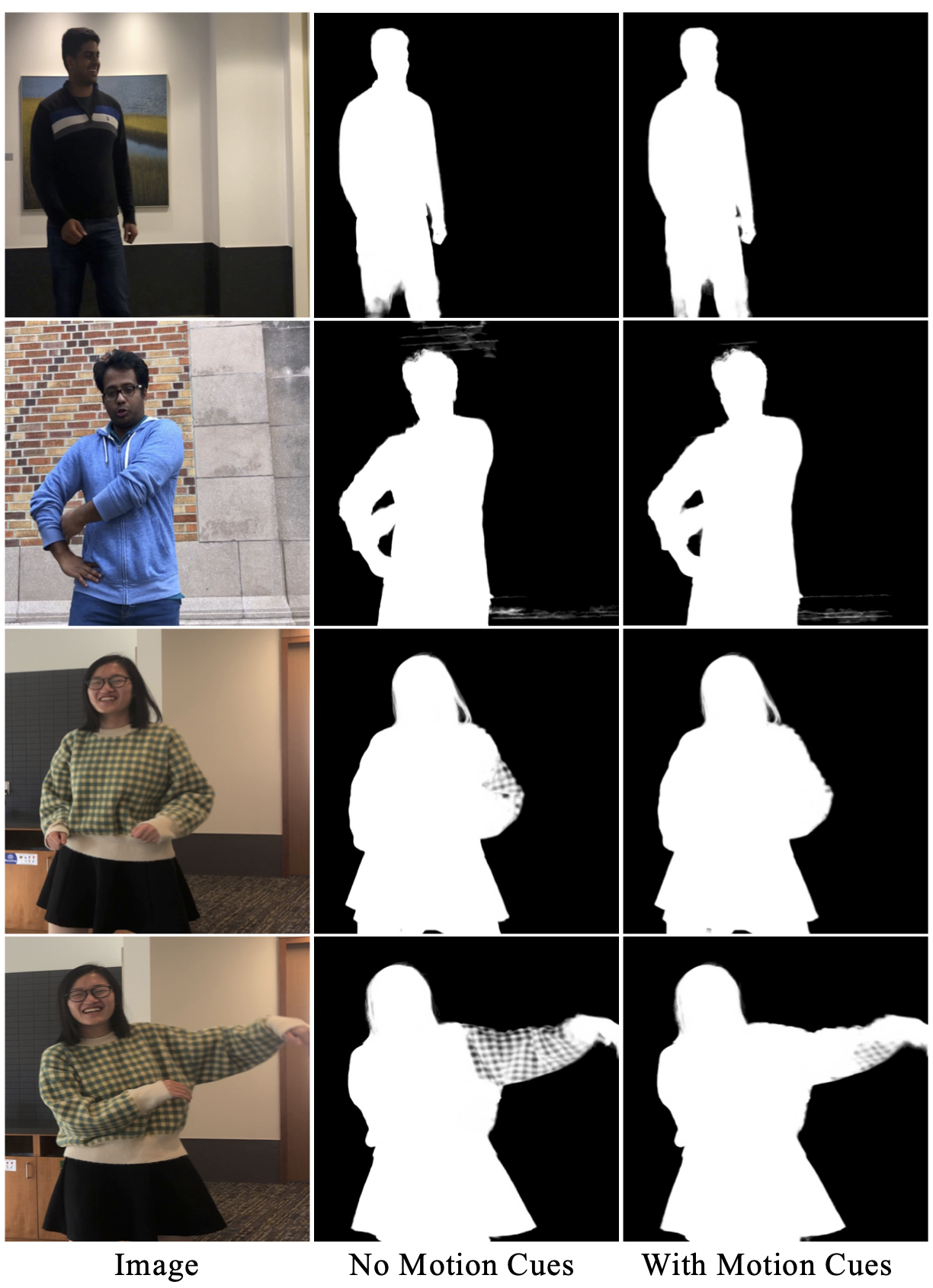}
	\caption{\small \textbf{Role of Motion Cues.} Motion cue helps in predicting better matte when foreground color coincides with the background and foreground moves in front of the background.}
	\label{fig:motion_cue_ab}	
\end{figure}

\subsection{Role of Motion Cues}
\label{sec:motion_cue}

Here we provide more detail and examples on the paper's ablation study for motion cues.  When the input is video, we have the option of setting the motion cue to $M=\{I_{-2T},I_{-T},I_{+T},I_{+2T}\}$, where T=20 for a 60fps video. We train another network $G_{\rm still}$, by removing the motion cue input and its related `Prior Encoder' and `Selector' from the architecture presented in Figure~\ref{fig:networks_main}. Thus $G_{\rm still}$ is the same as $G_{\rm Adobe}$, but without the motion cue block. We train both $G_{\rm Adobe}$ and $G_{\rm still}$ on the synthetic-composite Adobe dataset, following the same training protocol as described in the paper, for both networks. We then test both $G_{\rm Adobe}$ and $G_{concat}$ on our real video dataset.

In Figure~\ref{fig:motion_cue_ab} of this document and in Figure~5 of the main paper, we show the role motion cue plays in improving the alpha matte estimation. Specifically, the motion cue helps when foreground color matches the background, and when the foreground is moving significantly. Due to the foreground motion, the network can utilize additional frames (4 in this case) to determine that the regions which move are more likely to be foreground than the background, even though the color matches with the background. Note that, this may not be always true, e.g. a shadow cast on the background also moves with the foreground. Small camera motion with a handheld camera can also effectively cause motion in the background due to misregistration. Additionally, since we consider only a small time window of 1.33 secs for the motion cue, often there is lack of information as the foreground appears to be almost static during that time. 

To reiterate: whenever comparing to competing methods, we set the motion cue to $M=\{I,I,I,I\}$ and treat all images (including video frames) independently.


\begin{figure*}[!ht]
	\centering
	
	\begin{subfigure}[b]{1\textwidth}
  \includegraphics[clip,width=0.98\textwidth]{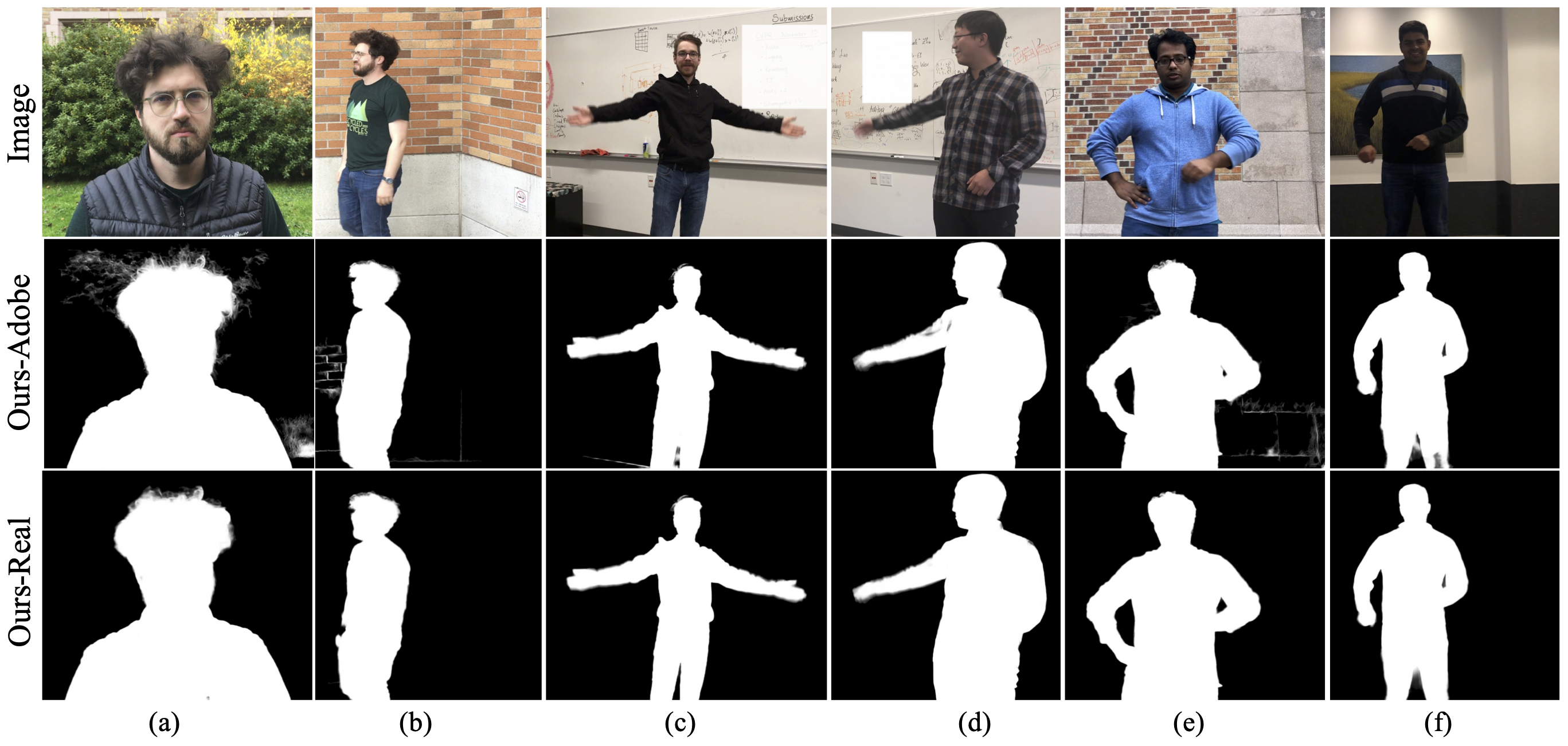}%
\end{subfigure}
\qquad
\begin{subfigure}[b]{1\textwidth}%
  \includegraphics[clip,width=0.98\textwidth]{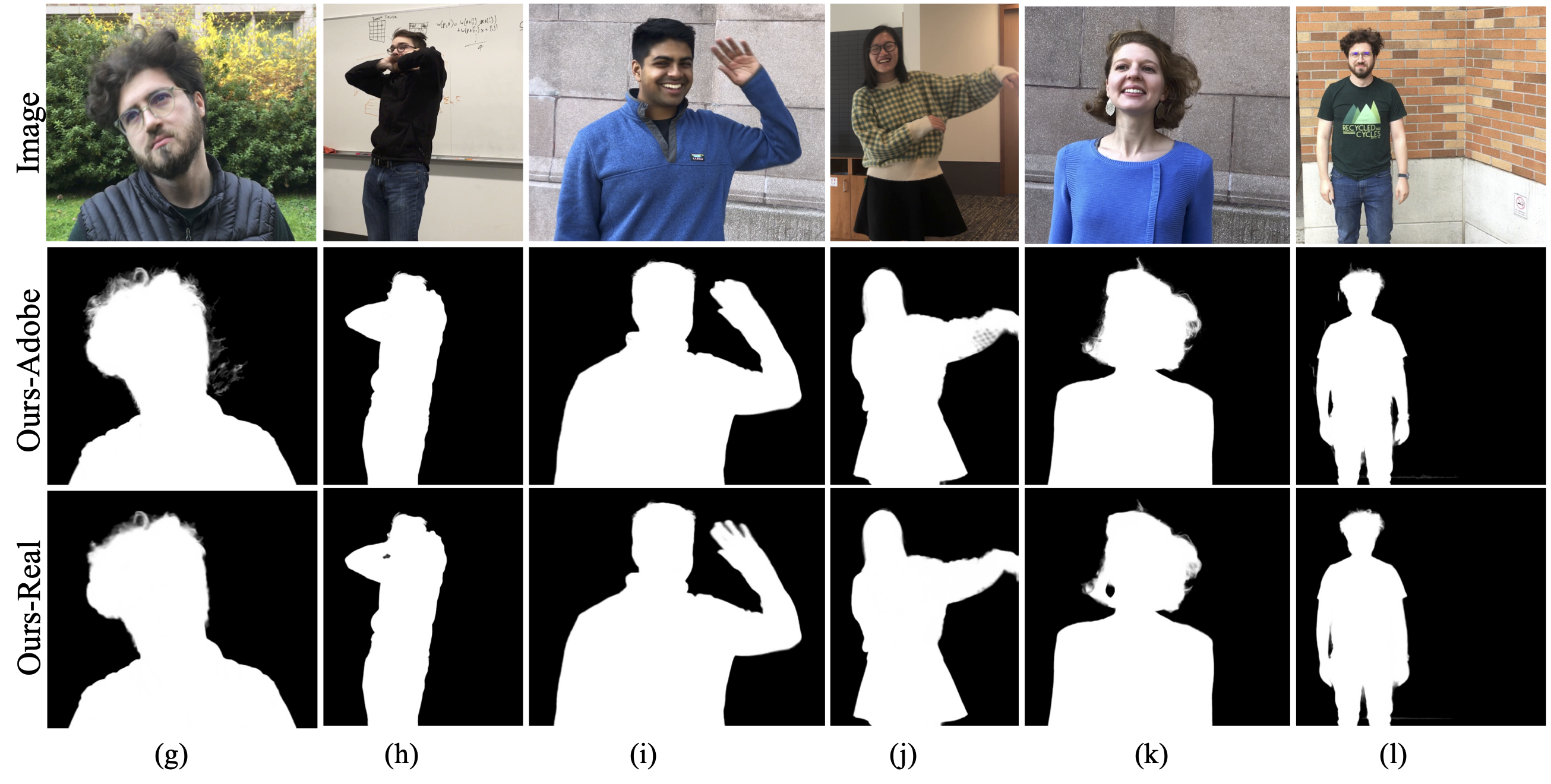}%
\end{subfigure}

	\caption{\small \textbf{Ours-Real vs Ours-Adobe.} `Ours-Real' is trained on real data guided by `Ours-Adobe' (trained on synthetic-composite dataset) along with an adversarial loss. (k) and (l) are instances where `Ours-Real' produces worse result compared to `Ours-Adobe'.}
	\label{fig:real_adobe_ab}	
\end{figure*}

\subsection{`Ours-Real' vs `Ours-Adobe'}
\label{sec:real_vs_syn}

Here we show more comparisons between using just the $G{\rm Adobe}$ network for matting (`Ours-Adobe') and using the full network $G_{\rm Real}$ guided by $G_{\rm Adobe}$ and discriminator $D$ (`Ours-Real').  In the main paper, we present an ablation study comparing `Ours-Real' with `Ours-Adobe' as both a user study in Table 3 and with qualitatively comparisons in Figure 6. Additional visual comparison on our test videos are presented in our project webpage. In Figure~\ref{fig:real_adobe_ab} of this document, we provide additional qualitative comparison between `Ours-Real' and `Ours-Adobe'. We find that `Ours-Real' is generally better though on occasion it is not; (k) and (l) are instances where `Ours-Real' produces an inferior matte compared to `Ours-Adobe'.

\subsection{Role of background and segmentation}
\label{sec:back_seg}

The captured background image without the subject and the estimated soft segmentation map are two key additional inputs that helps in estimating the foreground and alpha matte. We found that omitting the background image from the baseline $G_{Adobe}$ model degrades results substantially: SAD error of 8.33 without background vs. 1.73 with background on synthetic-composite Adobe dataset. For backgrounds that are relatively distant or roughly planar, homography-based alignment is accurate, and the network learns to handle remaining mis-registrations by training on hand-held videos. Alignment fails when the background, e.g., has two planes (Fig 4f, with two orthogonal walls).

Soft segmentation is obtained by eroding and dilating the segmentation predicted by Deeplabv3+. We observe that eroding and dilating the segmentation by 20 steps only increase the SAD error from 1.73 to 1.76 on synthetic-composite Adobe dataset. Hence our method is quite robust to errors in segmentation, and soft segmentation only indicates which is the foreground subject in the image. In contrast, the captured background plays more crucial role in the performance of the method.



\fi

\end{document}